\documentclass[letterpaper, 10 pt, journal, twoside]{IEEEtran}

\usepackage{graphics} 
\usepackage{amsmath} 
\usepackage{amssymb}  
\usepackage{bm}
\usepackage{cleveref} 
\usepackage{paralist} 
\usepackage[dvipsnames,table,xcdraw]{xcolor}
\usepackage{multirow}
\usepackage{graphicx}
\usepackage{url}
\usepackage{dblfloatfix}
\usepackage{booktabs}
\usepackage{epstopdf}
\usepackage{soulutf8}
\soulregister{\cite}7 
\soulregister{\eqref}7 %
\soulregister{\ref}7 %

\usepackage[backend=biber,      
    style=ieee,
    bibstyle=ieee,              
    sortcites=true,   
    mincitenames=1,
    maxcitenames=2,
    giveninits=true,            
    backref=false
]{biblatex}

\usepackage{caption}
\usepackage{atbegshi}

\newcommand{\Real}{\mathbb{R}}

\newcommand{\x}{\bm{x}}
\renewcommand{\v}{\bm{v}}

\newcommand{\q}{\bm{q}}
\newcommand{\dq}{\dot{\bm{q}}}

\newcommand{\dv}{\dot{\bm{v}}}
\newcommand{\F}{\bm{F}}

\renewcommand{\u}{\bm{u}}

\newcommand{\X}{\bm{\mathcal{X}}}

\newcommand\norm[1]{\left\lVert#1\right\rVert}
\newcommand{\new}[1]{#1} 

\newcommand{\newtwo}[1]{#1}

\begin{document}

\AtBeginShipout{\AtBeginShipoutUpperLeft{%
  \put(\dimexpr\paperwidth-3.9cm\relax,-0.7cm){\makebox[0pt][r]{\framebox{\footnotesize This work is published in IEEE Robotics and Automation Letters, DOI :https://doi.org/10.1109/LRA.2024.3455907 }}}%
}}

\title{
Tailoring Solution Accuracy for Fast Whole-body Model Predictive Control of Legged Robots
}

\author{Charles Khazoom$^{1}$, Seungwoo Hong$^{1}$, Matthew Chignoli$^{1}$, Elijah Stanger-Jones$^{1}$ and Sangbae Kim$^{1}$
\thanks{Manuscript received: June, 29, 2024; Revised August, 15, 2024; Accepted September, 2, 2024.}
\thanks{This paper was recommended for publication by Editor A. Kheddar upon evaluation of the Associate Editor and Reviewers' comments.
This work was supported by Naver Labs, NSERC and FRQNT.}
\thanks{$^{1}$Authors are with Department of Mechanical Engineering Department,
        Massachusetts Institute of Technology,  77 Massachusetts Ave, Cambridge, MA, United States
        {\tt\footnotesize ckhaz@mit.edu, sangbae@mit.edu}}%
\thanks{Digital Object Identifier (DOI): see top of this page.}
}


\maketitle

\begin{abstract}
    Thanks to recent advancements in accelerating non-linear model predictive control (NMPC), it is now feasible to deploy whole-body NMPC at real-time rates for humanoid robots.
However, enforcing inequality constraints in real time for such high-dimensional systems remains challenging due to the need for additional iterations.
This paper presents an implementation of whole-body NMPC for legged robots that provides low-accuracy solutions to NMPC with general equality and inequality constraints.
Instead of aiming for highly accurate optimal solutions, we leverage the alternating direction method of multipliers to rapidly provide low-accuracy solutions to quadratic programming subproblems.
Our extensive simulation results indicate that real robots often cannot benefit from highly accurate solutions due to dynamics discretization errors, inertial modeling errors and delays.
We incorporate control barrier functions (CBFs) at the initial timestep of the NMPC for the self-collision constraints, resulting in up to a 26-fold reduction in the number of self-collisions without adding computational burden.
The controller is reliably deployed on hardware at 90 Hz for a problem involving 32 timesteps, 2004 variables, and 3768 constraints.
The NMPC delivers sufficiently accurate solutions, enabling the MIT Humanoid to plan complex crossed-leg and arm motions that enhance stability when walking and recovering from significant disturbances.
\end{abstract}

\IEEEpeerreviewmaketitle
\section{Introduction}
\label{sec:introduction}
\IEEEPARstart{L}{egged} robots have great potential for applications requiring animal-like mobility, but their high dimensionality, non-linearity, and underactuation make them difficult to control.
\par
Reinforcement learning (RL) policies showed promise in addressing these challenges, partly because they are trained in mature high-fidelity simulators with whole-body dynamics and kinematics.
However, RL requires vast amounts of data \new{and cannot easily enforce system constraints}.
\par
Non-linear model predictive control (NMPC) is another powerful tool for providing closed-loop stability to constrained non-linear systems. Unlike RL, NMPC uses a model to rapidly design new behaviors via online optimization.
Yet, achieving real-time rates for high-dimensional systems like legged robots is challenging, especially when aiming for highly accurate optimal solutions with inequality constraints.
However, real robotic systems may not always benefit from high-accuracy solutions due to modeling errors and delays.
\par
\begin{figure}[t]
    \centering
    \vspace{3mm}
    \includegraphics[scale=0.95]{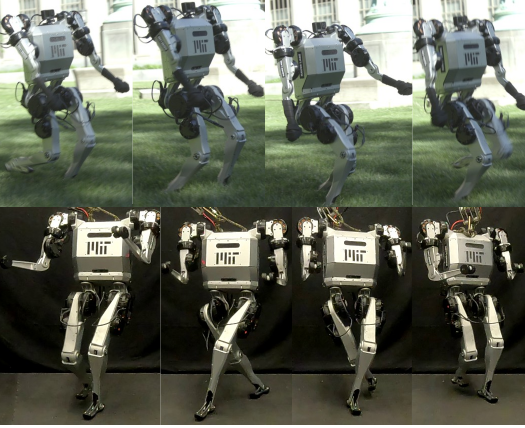}
    \caption{Hardware demonstration of the whole-body NMPC on the MIT Humanoid.
    Top row: walking with emergent arm swing.
    Bottom row: leg crossing with dynamic self-collision avoidance with right leg fully extended during abrupt halting during lateral walking.}
    \captionsetup{justification=raggedright, singlelinecheck=false}
    \caption*{\new{Supplementary video :} \url{https://youtu.be/Xmi7AFGhlFc}}
    \label{fig:hardware_figures}
\end{figure}

We present an implementation of whole-body NMPC with inequality constraints.
Using a suitable solver, we tailor solution accuracy to the computational budget and improve self-collision avoidance using control barrier functions (CBFs) at the first timestep of the NMPC.
We demonstrate our approach through extensive simulation and hardware walking experiments on the MIT Humanoid (Fig. \ref{fig:hardware_figures}).

\subsection{Multilayered Controller Architectures for Legged Robots}
To reduce the computational burden of NMPC for legged robots, the problem is usually broken down into smaller multilayered optimizations that are easier to solve.
A common architecture uses MPC with reduced-order models, whose outputs are tracked by an inverse dynamics whole-body controller (WBC) \cite{herdtOnlineWalkingMotion2010, kimHighlyDynamicQuadruped2019}.
These simple models enable fast computation, but cannot account for all degrees of freedom and constraints like joint position limits, torque limits and self-collisions, which require holistic whole-body reasoning.
\par
This issue can be partly adressed by reactive WBCs which project desired accelerations on the constraints of a whole-body model.
The constraints are typically limited to ones at the acceleration level like friction cone or torque constraints, but non-linear constraints on the robot's configuration can be incorporated via control barrier functions (CBFs) \cite{grandiaMultiLayeredSafetyLegged2021, grandiaPerceptiveLocomotionNonlinear2023}.
For instance, exponential CBFs \cite{khazoomHumanoidSelfCollisionAvoidance2022} were previously incorporated in a WBC for self-collision avoidance.
These reactive WBCs benefit from exact linearization at the current state, resulting in small quadratic programs solvable at high frequency by off-the-shelf solvers.
However, they are myopic to the future.
Additionally, multilayered architectures can produce conflicts between the reduced-order MPC and the WBC since they use different models, horizons and constraints.
Ideally, we would need a single planner that incorporates all the robot's degrees of freedom and constraints and provides real-time solutions for long-horizon tasks.
However, incorporating whole-body dynamics, kinematics and arbitrary constraints within a single NMPC controller is still computationally challenging for high-dimensional systems.

\subsection{Solution Methods for Real-Time NMPC of Legged Robots}
Various strategies are used to address the computational burden of whole-body NMPC.
Common solution methods include sequential quadratic programming (SQP) \cite{nocedalNumericalOptimization2006} and differential dynamic programming (DDP) \cite{mayneSecondorderGradientMethod1966}, both of which solve a local convex subproblem at each iteration.
Solving NMPC to optimality can be prohibitively slow; therefore, only one or a few iterations are typically performed online at each control instant \cite{diehlRealTimeIterationScheme2005}.
Furthermore, specialized matrix factorization methods that exploit the temporal structure of the problem (e.g., Riccati recursion), specialized algorithms for analytical derivatives \cite{carpentierAnalyticalDerivativesRigid2018, katayamaEfficientSolutionMethod2021,katayamaModelPredictiveControl2023} and parallelization \cite{jalletParallelProximalLinearQuadratic2024, katayamaEfficientSolutionMethod2021} enabled real-time NMPC on hardware for quadrupeds \cite{neunertWholeBodyNonlinearModel2018,sleimanUnifiedMPCFramework2021, grandiaPerceptiveLocomotionNonlinear2023, mastalliInverseDynamicsMPCNullspace2023} and humanoids \cite{koenemannWholebodyModelpredictiveControl2015, dantecWholeBodyModelPredictive2022}.
\par
Despite these advancements, inequality constraints are still hard to enforce in real time for high-dimensional systems since they require additional expensive inner iterations to solve the local subproblems. 
For instance, interior point solvers \cite{frisonHPIPMHighperformanceQuadratic2020, schwanPIQPProximalInteriorPoint2023a} require multiple matrix factorizations and active set methods scale poorly with the number of inequalities \cite{ nocedalNumericalOptimization2006}.
Augmented Lagrangian approaches have increased in popularity for handling constraints for QP \cite{bambadePROXQPEfficientVersatile, schwanPIQPProximalInteriorPoint2023a} and trajectory optimization solvers \cite{jalletPROXDDPProximalConstrained, howellALTROFastSolver2019}.
Yet, even the most advanced implementations of whole-body NMPC for legged robots relax inequality constraints like friction cone, joint limits and self-collisions in the cost function using penalty methods \cite{chiuCollisionFreeMPCWholeBody2022, grandiaPerceptiveLocomotionNonlinear2023, jalletPROXDDPProximalConstrained}.
This technique increases the coupling between cost minimization and constraint satisfaction \cite{sleimanConstraintHandlingContinuousTime2021}, leading to time-consuming and task-specific tuning.
\textcite{sleimanConstraintHandlingContinuousTime2021} proposed an online scheme based on the method of multipliers to handle inequality constraints but can only afford a single update of the decision variables and Lagrange multipliers per control instant \cite{sleimanConstraintHandlingContinuousTime2021,sleimanUnifiedMPCFramework2021}.
\par
Another augmented Lagrangian technique for solving QP subproblems is the alternating direction method of multipliers (ADMM) \cite{boydDistributedOptimizationStatistical2010}.
ADMM is a general framework that splits optimization problems into smaller subproblems that are easier to solve and has gained popularity in robotics \cite{budhirajaDynamicsConsensusCentroidal2019,
meduriBiConMPNonlinearModel2023, bravo-palaciosLargeScaleADMMbasedCoDesign2022, aydinogluRealTimeMultiContactModel2022}.
In particular, the operator splitting QP solver OSQP leverages ADMM to handle linear equality and inequality constraints while only requiring a single sparse matrix factorization followed by multiple cheap iterations \cite{stellatoOSQPOperatorSplitting2020}.
Yet, there is currently no demonstration of its applicability for solving the QP subproblems arising in whole-body NMPC of legged robots in real time, even though some SQP implementations using OSQP exist \cite{anderssonCasADiSoftwareFramework2019, verschuerenAcadosModularOpensource2022} and have been deployed on lower-dimensional robot manipulators \cite{jordanaStagewiseImplementationsSequential2023}.
While ADMM may need many iterations to provide high-accuracy solutions, it can rapidly converge to low-accuracy solutions and be warm-started, making it a practical option for solving very complex NMPC problems in real time.
\par
\newtwo{In this paper, we present a whole-body NMPC formulation and a solver that leverages ADMM to solve the inner QP subproblem of an SQP method with inequality constraints in real time.}

\subsection{Contributions}
The main contributions of this paper are threefold.
\newtwo{First, we demonstrate the advantages of rapid, low-accuracy QP subproblem solutions obtained from ADMM by benchmarking the impact of solution accuracy on closed-loop performance in the presence of unavoidable inertial modeling errors, dynamics discretization errors, and computation delay.}
\par
Second, we enhance the closed-loop feasibility of low-accuracy solutions by incorporating CBF constraints at the first timestep of the NMPC.
These constraints are easier to satisfy due to exact linearization at the measured state, making them independent of the initial guess.
This approach enables stricter self-collision avoidance within the NMPC with minimal added complexity while avoiding conflicts that arise in multilayered controllers.
\par
Lastly, we validate our approach through comprehensive hardware experiments on the MIT Humanoid.
These experiments demonstrate the emergence of coordinated whole-body movements, including stabilizing arm and crossed-leg motions.
The robot reliably respects joint limits, self-collisions and contact constraints while walking and recovering from significant external disturbances.
\par
The paper is organized as follows.
Section \ref{sec:controller_design} presents the NMPC formulation and the SQP method to solve it.
Section \ref{sec:sim_results} presents the simulation analyses.
Section \ref{sec:hardware} presents the hardware demonstration on the MIT Humanoid.
\section{Controller Design}
\label{sec:controller_design}

This section presents background on NMPC, details our proposed NMPC formulation, and the solution strategy for solving it in real time.
\subsection{Background on Non-linear MPC}
NMPC closes the loop by continuously solving the non-linear trajectory optimization problem \eqref{eq:NMPC} with horizon $N \in \mathbb{N}$ and applying the first control input $\u_0$ as the state of the robot evolves:
\begin{subequations}
\label{eq:NMPC}
\begin{align}
    \min_{\X} &\sum_{k=0}^{N-1}l(\x_k,\u_k, \bm{\theta}) + l_N(\x_N,\bm{\theta}) \\  
             \mathrm{subject~to} \nonumber\\ 
            \x&_{0} = \hat{\x}\\ 
            \x&_{k+1} = \bm{f}(\x_k,\u_k, \bm{\theta}) ~\forall~k \in \{0,...,N-1\}\\
            \bm{g}&(\x_k,\u_k, \bm{\theta})\le\bm{0} ~\forall~k \in \{0,...,N-1\}\\
            \bm{g}&_N(\x_N, \bm{\theta})\le\bm{0},
\end{align}
\end{subequations}
where ${\X}$ is the set of decision variables composed of the states $\x_k$ and control inputs $\u_k$ at stage $k \in \{0,...N\}$, $\hat{\x}$ is the measured state, $\bm{\theta}$ is a parameter vector, $l(\cdot)$ is the running cost, $l_N(\cdot)$ is the terminal cost, $\bm{f}(\cdot)$ is the discrete-time dynamics, $\bm{g}(\cdot)$ and $\bm{g}_N(\cdot)$ are the path equality and inequality constraints.

\subsection{Whole-body Non-Linear MPC Formulation}
\label{sec:NMPC_formulation}
\subsubsection{Decision Variables}
The whole-body NMPC solves for the set of decision variables $\X$:
\begin{equation}
\X := \left\{\left\{\q_k, \v_k\right\}_{k=0}^{N}, \left\{\bm{\tau}_k,\left\{\F_{c,k}\right\}_{\forall c \in \mathcal{C}}\right\}_{k=0}^{N-1}\right\} .
\end{equation}
For a robot with $n_j$ actuated joints, $\X$ includes the robot's configuration $\q \in \Real^{n_j+6}$, generalized velocity $\v \in \Real^{n_j+6}$, the joint torques $\bm{\tau} \in \Real^{n_j}$ and the contact forces $\bm{F}_c \in \Real^{3}$ for each contact point $c \in \mathcal{C}$.
The configuration $\q$ is composed of the floating base position $\bm{p} \in \Real^3$, Euler angle orientation $\bm{\Theta} \in \Real^3$, and joint position $\q_j \in \Real^{n_j}$.
The velocity $\v$ includes the spatial velocity of the floating base $\v_{b} \in \Real^{6}$ and the joint velocity $\dq_j \in \Real^{n_j}$.

\subsubsection{\new{Cost Function}}
\new{The running cost is given by} 
\begin{gather}
    l = \norm{\delta\v_k}_{\bm{W}_{\v}}^2 + \norm{\delta\q_k}_{\bm{W}_{\q}}^2 + \textstyle \sum_{c\in\mathcal{C}}\norm{\delta\F_{c,k}}_{\bm{W}_{\F_c}}^2,
    \label{eq:cost_fcn}
\end{gather}
\new{where $\delta \cdot$ indicates the difference between a variable and its reference, $\norm{\cdot}$ is the L2-norm, and $\bm{W}_{\q}$, $\bm{W}_{\v}$, $\bm{W}_{\F_c}$ are diagonal weight matrices for the corresponding variable.
In the first term of \eqref{eq:cost_fcn}, the torso velocity command is non-zero in the forward, lateral and yaw directions, and zero otherwise.
The joint velocity command is zero.
In the second term of \eqref{eq:cost_fcn}, the torso position reference is integrated from the velocity command (constant along the NMPC horizon), with constant height.
The joint position command is a fixed standing pose.
The reference for $\F_{c,k}$ in the vertical direction is the weight of the robot distributed across the contact points and zero otherwise.
For the first timestep, a regularization term $\norm{\bm{\tau}_0 - \bm{\tau}_{\mathrm{prev}}}_{\bm{W}_{\bm{\tau}_0}}^2$ weighted by diagonal matrix ${\bm{W}_{\bm{\tau}_0}}$ eliminates hardware vibrations by penalizing large deviations between $\bm{\tau}_0$ and the previous torque $\bm{\tau}_{\mathrm{prev}}$.}
The cost function has a least-squares form, and the Hessian of the Lagrangian is approximated using the Gauss-Newton Hessian to reduce solve time \cite{diehlRealTimeIterationScheme2005}.

\subsubsection{Dynamics Constraints}
\label{sec:dynamics_constraints}
The discrete-time dynamics are described as:
\begin{gather}
    \q_{k + 1} = \q_k + \frac{1}{2}\bm{B}(\bm{\Theta_k})(\v_{k + 1} + \v_k) dt_k, \label{eq:int_constr_pos}
\end{gather}
where $dt_k$ is the dynamics discretization timestep and $\bm{B}(\bm{\Theta})$ maps $\v$ to $\dq$.
\par
The trajectory is constrained by the inverse dynamics $\bm{f}_\mathrm{RNEA}(\cdot)$, efficiently computed with the recursive Newton-Euler algorithm (RNEA):
\begin{gather}
    \begin{bmatrix}
        \bm{0}_{6\times 1}\\
        \bm{\tau}_k
    \end{bmatrix} = \bm{f}_{\mathrm{RNEA}}(\q_k, \v_k,\dv_k) - \sum_{c \in \mathcal{C}}\bm{J}_{c,k}^\top \F_{c,k} \label{eq:rnea_fb},
\end{gather}
where $\bm{J}_c \in \Real^{3\times (n_j+6)}$ is the Jacobian of contact point $c$.
\new{The RNEA algorithm and its derivatives are computationally cheaper to compute than the forward dynamics, and can advantageously increase the sparsity of the NMPC \cite{katayamaEfficientSolutionMethod2021}}.
The wrench on the floating base is zeroed, and $\bm{\tau}_k$ corresponds to the last $n_j$ rows in \eqref{eq:rnea_fb}.
The acceleration $\dv_k$ is computed from finite differences:
\begin{equation}
    \dv_k = \frac{\v_{k + 1} - \v_k}{dt_k}\label{eq:qdd_finite_diff}.
\end{equation}
\subsubsection{Contact and Swing Constraints}
\label{sec:contact_swing_constraints}
The NMPC takes as input a contact sequence $\gamma_{c,k}\in \left\{0,1\right\}$ for each contact point $c$ and stage $k$, where  $\gamma_{c,k} = 1$ in contact and $\gamma_{c,k} = 0$ in swing.
Contact forces are constrained to $\bm{0}_{3\times1}$ in swing and to the friction cone in stance: 
\begin{gather}
    (1-\gamma_{c,k})\F_{c,k} = \bm{0}_{3\times1}\\
    \gamma_{c,k}\mu F_{c,z,k} \ge \gamma_{c,k}\sqrt{F_{c,x,k}^2 + F_{c,y,k}^2},
\end{gather}
where $F_{c,x,k}$ and $F_{c,y,k}$ are the world-frame tangential forces in the forward and lateral directions.
The multiplication by \((1 - \gamma_{c,k}\phantom{})\)  and $\gamma_{c,k}$ respectively deactivates the constraints in stance and swing without removing them from the problem at runtime.
This keeps the sparsity structure of the QP subproblems fixed, which facilitates warmstart and avoids repeating expensive symbolic factorizations.
\par
The heights of the contact points $f_{c,z}(\q_k)$ are constrained to follow a reference height $z_{c,k}$:
\begin{gather}
    f_{c,z}(\q_k) = z_{c,k} \label{eq:foot_height}.
\end{gather}
Similarly, the tangential velocity of the contact points is zeroed in stance: 
\begin{gather}
    \gamma_{c,k}\bm{J}_{c,xy,k}\dq_k = \bm{0}_{2\times1}\label{eq:no_slip},
\end{gather}
where $\bm{J}_{c,xy,k}$ maps $\dq_k$ to the velocity tangent to the ground.
We use the adaptive segmentation algorithm detailed in \cite{bledtImplementingRegularizedPredictive2019} to segment each $\gamma_{c,k}$ and $dt_k$ \new{such that the predicted gait events (transitions from stance to swing and from swing to stance) match the predicted gait along the prediction horizon.}
\par
\subsubsection{Box and Kinematic Inequality Constraints}
\begin{figure}[!htbp]
    \centering
    \includegraphics[scale=1]{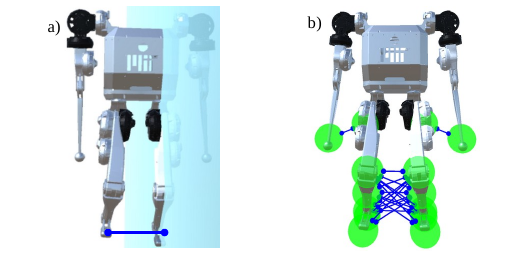}
    \caption{Two self-collision models. a) the simple collision model constrains the right foot to a minimum distance from a plane attached to the left foot. b) the complex collision model composed of 19 pairs of spheres. Each blue line corresponds to a distance constraint between a pair of spheres.}
    \label{fig:self_collision_constraint}
\end{figure}

Simple box constraints are applied to $\q_j$, $\dq_j$, and $\bm{\tau}$ to respect joint position, velocity and torque limits.
To avoid self-collisions, non-linear kinematic inequality constraints are enforced:
\begin{equation}
    \bm{h}(\q_k) \ge \bm{0}. \label{eq:kin_constraint}
\end{equation}
Equation \eqref{eq:kin_constraint} can be simple distance constraints that prevent the right foot from crossing a virtual plane attached to the left foot (Fig. \ref{fig:self_collision_constraint}a), or the signed-distance functions between spheres (Fig. \ref{fig:self_collision_constraint}b).
We used the former in Section \ref{sec:sim_results} to simplify analysis and the latter on the real robot to plan more complex motions with crossed legs.

\subsubsection{Control Barrier Functions}
The self-collision constraints \eqref{eq:kin_constraint} are highly non-linear and challenging to satisfy with less than one-centimeter accuracy when the robot experiences significant disturbances, especially in the real-time setting where a single SQP iteration is performed at each control instant.
Another option to strictly enforce \eqref{eq:kin_constraint} is to use exponential CBFs \cite{nguyenExponentialControlBarrier2016, khazoomHumanoidSelfCollisionAvoidance2022, grandiaMultiLayeredSafetyLegged2021}, which constrain $\dv$ to ensure future satisfaction of \eqref{eq:kin_constraint}.
The CBF constraint is
\begin{equation}
\bm{J_h}\bm{\dv} + \bm{\dot{J}_h}\bm{\dq} + (\alpha_1+\alpha_2) \bm{J_h}\dq + (\alpha_1\alpha_2)\bm{h}(\q) \ge \bm{0}, \label{eq:cbf_constraint}
\end{equation}
where $\bm{J_h}(\q)$ is the Jacobian of $\bm{h}(\q)$ with respect to $\q$, and $\alpha_1, \alpha_2$ are scalars that bound the time-response of $\bm{h}(\q)$.
This constraint has proven effective for reactive self-collision avoidance in humanoid robots \cite{khazoomHumanoidSelfCollisionAvoidance2022} and relates to the CBF-QP safety filter \cite{amesControlBarrierFunctions2019}.
\par
When applied at the first timestep, \eqref{eq:cbf_constraint} is evaluated at the measured state, making it linear and independent of the initial guess.
These properties make it less susceptible to solver inaccuracies arising from performing a single SQP iteration and early termination of the ADMM QP solver.
Thus, we apply \eqref{eq:cbf_constraint} at the first timestep and \eqref{eq:kin_constraint} at the remaining timesteps.
The former is more conservative but easier to solve, while the latter allows for planning more complex motions.

\subsection{Solution Method}
\newtwo{The NMPC formulation is solved with a custom general-purpose solver that exploits sparsity.} Alternatively, a solver tailored for optimal control, as in \cite{jordanaStagewiseImplementationsSequential2023} could be used by defining $\x_k = [\q_k, \v_k ]^{\top}$ and $\u_k = [\bm{\tau}_k,  \{\F_{c,k}\}_{\forall c \in \mathcal{C}}, \v_{k+1}]^{\top}$, \new{and recasting the problem into the form of \eqref{eq:NMPC}.}
\par
The NMPC is expressed in the form of the general non-linear program \eqref{eq:nlp}
\begin{equation}
\label{eq:nlp}
\begin{aligned}
\underset{\bm{z}}{\text{min}}
~~~&C(\bm{z}, \bm{\theta}) \\
\text{s.t.}~~~
 \bm{L}(\bm{\theta}) \le & ~\bm{G}(\bm{z}, \bm{\theta}) \le \bm{U}(\bm{\theta}),
\end{aligned}
\end{equation}
where $\bm{z}$ is the decision variable vector, $\bm{\theta}$ is the parameter vector, $C(\cdot)$ is the cost function, $\bm{G}(\cdot)$ is the vector of constraints with lower and upper bounds $\bm{L}(\cdot)$ and $\bm{U}(\cdot)$.
\par
One SQP step with the Gauss-Newton Hessian approximation consists in solving the following local QP subproblem at each iteration $i$:
\begin{equation}
\begin{aligned}
\label{eq:sqp_step}
&\underset{\bm{d}_i}{\text{min}}
~~~\frac{1}{2} \bm{\bm{d}_i}^T \nabla^2_{z} C(\bm{z}_{i}, \bm{\theta}) \bm{\bm{d}_i} + \nabla_{\bm{z}} C(\bm{z}_{i}, \bm{\theta})^T \bm{d}_i\\
&\text{s.t.}~~~
\bm{L}(\bm{\theta}) \leq \bm{G}(\bm{z}_{i}, \bm{\theta}) + \nabla_{\bm{z}} \bm{G}(\bm{z}_{i}, \bm{\theta}) \bm{d}_i \leq \bm{U}(\bm{\theta}),
\end{aligned}
\end{equation}
where $\bm{d}_i$ is the step direction, $\nabla_{\bm{z}} C^{\top}$ is the gradient of $C(\cdot)$ with respect to $\bm{z}$, $\nabla_{\bm{z}} \bm{G}$ is the Jacobian of $\bm{G}(\cdot)$, and $\nabla^2_{z} C$ is the Hessian of the quadratic cost $C(\cdot)$ and corresponds to the Gauss-Newton Hessian.
The solution is updated as follows: 
\begin{equation}    
\bm{z}_{i+1} = \bm{z}_{i} + \beta_i\bm{d}_i,
\end{equation}
where $\beta_i$ is obtained from a linesearch algorithm.
\par
Solving \eqref{eq:sqp_step} is the most computationally expensive step, especially when inequality constraints are present.
In this work, we solve \eqref{eq:sqp_step} using the ADMM algorithm from the QP solver OSQP \cite{stellatoOSQPOperatorSplitting2020}, which provides low-accuracy solutions within a few cheap iterations and can easily be wamstarted.
This allows to tailor the solution accuracy to the computational budget by running a small but fixed number of iterations at high rates.

\subsection{Implementation Details}
The controller architecture for the MIT Humanoid is composed of the NMPC, whose trajectories are interpolated at 1~kHz and fed in the motor-level PD controller running at 20~kHz.
\subsubsection{The MIT Humanoid}
The MIT Humanoid \cite{saloutosDesignDevelopmentMIT2023} stands 1.04~m tall and weighs 24~kg.
It has 24 degrees of freedom, and each leg and arm have five and four actuated joints, respectively ($n_j = 18$).
The feet don't have ankle roll, making dynamic locomotion necessary since there is no support polygon when in single-leg stance.
A parallel belt transmission actuates the knee and ankle joints.
This couples the knee and ankle but allows the motors to be closer to the torso.
All computations run on an onboard UP Xtreme i12 computer with a Core i7-1270PE CPU.
\subsubsection{NMPC Implementation for the MIT Humanoid}
The robot's spinning rotors can contribute significantly to its dynamics and are therefore included in the dynamics model.
The particularly challenging effects of the coupled knee and ankle rotors are handled via the constraint embedding-based algorithms \cite{jainRecursiveAlgorithmsUsing} implemented in \cite{chignoliRecursiveRigidBodyDynamics2023}.
The NMPC formulation uses two contact points per foot, i.e. $\mathcal{C}=\{\mathrm{Right~Heel,~Right~Toe,~Left~Heel,~Left~Toe}\}$.
The gait sequence is specified individually for each contact point to enable smoother heel-to-toe and toe-to-heel gaits.
Because the MIT Humanoid has been designed for high-power jumping tasks \cite{saloutosDesignDevelopmentMIT2023}, walking tasks are not torque-limited.
Thus, the joint torque box constraint and the last $n_j$ rows of constraint \eqref{eq:rnea_fb} are only applied for the first two timesteps to reduce the computation time.
The constraints and derivatives required by \eqref{eq:sqp_step} are computed using CasADi's automatic differentiation and codegeneration tools \cite{anderssonCasADiSoftwareFramework2019}.
\par
We run a single SQP iteration at every control instant, \new{utilizing the previous unshifted solution as an initial guess.}
The inner OSQP solver runs a fixed number of iterations.
This allows to update solutions at deterministic rates and facilitates prediction delay compensation, which is performed by linearly interpolating the solved trajectory forward in time.
\new{The default sparse QDLDL linear solver is used to efficiently exploit the problem's temporal sparsity and the robot's kinematics and dynamics sparsity}.
\new{To reduce hardware vibrations and avoid additional time-consuming matrix factorizations, the} \verb|adaptive_rho| \new{option in OSQP is disabled.}
\new{The linesearch algorithm from \cite{grandiaPerceptiveLocomotionNonlinear2023} is used as it also helped reduce vibrations on hardware.}
\par
The whole-body NMPC is implemented in C++ and runs at 90-200~Hz on the robot's computer, depending on the formulation used.
All NMPC computations run on a single thread, as opposed to most other whole-body NMPC implementations relying on multithreading to reduce computation time \cite{koenemannWholebodyModelpredictiveControl2015,katayamaEfficientSolutionMethod2021, dantecWholeBodyModelPredictive2022, grandiaPerceptiveLocomotionNonlinear2023, jalletParallelProximalLinearQuadratic2024}.
\subsubsection{Low-level Control}
The joint positions, velocities and torques solved by the NMPC are interpolated at 1~kHz.
The resulting feedforward torque $\bm{\tau}_\mathrm{ff}$, position and velocity setpoints $\q_\mathrm{des}$, $ \dq_\mathrm{des}$ are fed in the motor-level controller, where the commanded motor torque $\bm{\tau}_\mathrm{m}$ is computed by the following PD control law at 20~kHz:
\begin{equation}
    \bm{\tau}_\mathrm{m} = \bm{\tau}_\mathrm{ff} + \bm{K}_\mathrm{p}(\q_\mathrm{des} - \q) + \bm{K}_\mathrm{d}(\dq_\mathrm{des} - \dq).
    \label{eq:pd_ctrl}
\end{equation}
The matrices $\bm{K}_\mathrm{p}$ and $\bm{K}_\mathrm{d}$ are diagonal matrices with elements set to 50~N$\cdot$m/rad and 2~N$\cdot$m$\cdot$s/rad, respectively.

\subsubsection{State Estimation}
\new{
A linear Kalman filter fuses contact state estimates, joint encoders and accelerometer measurements to estimate the linear velocity of the torso, as in \cite{bledtMITCheetahDesign2018}.
The contact states of the heels and toes are assumed to match the pre-specified gait.
This strategy was sufficient to walk stably on uneven terrain as shown in Section \ref{sec:hardware}}.

\section{Simulation Experiments}
\label{sec:sim_results}
The simulations aim to assess the effect of the solution accuracy of the QP subproblem \eqref{eq:sqp_step} on closed-loop performance considering three sources of error: inertial errors, dynamics discretization errors, and computation delay.
We also evaluate the impact of CBFs on self-collision avoidance.

\subsection{Experimental Setup}
The NMPC uses an approximate contact model with coarse dynamics integration timesteps, but is evaluated in a custom high-fidelity simulator \new{\cite{CheetahSoftware2024}}.
This simulator enforces contact complementarity constraints, friction cone, and actuator torque-speed limits with a fine timestep of $0.5~$ms.
\par
To evaluate a controller's closed-loop performance, we conduct large-scale simulations comprising 500 individual runs, each lasting 6 seconds.
During each simulation, the robot is initialized in the same standing pose. It starts walking while subjected to a constant velocity command $\bm{v}_\mathrm{cmd} = [v_{x,\mathrm{cmd}}, v_{y,\mathrm{cmd}}, \omega_{z,\mathrm{cmd}}] \in [-3,3]~\mathrm{m/s} \times [0,2]~\mathrm{m/s} \times [0,5]~\mathrm{rad/s}$, where $v_{x,\mathrm{cmd}}$, $v_{y,\mathrm{cmd}}$, and $\omega_{z,\mathrm{cmd}}$ are the forward, lateral, and turning velocity commands, respectively.
A simulation is deemed successful if the robot avoids falling and satisfies the simple self-collision constraints (Fig. \ref{fig:self_collision_constraint}a) throughout the entire duration.
\new{Every controller experiences the same set of 500 velocity commands selected randomly.}
\par
We repeat all 500 simulations for different levels of solution accuracy, inertial modeling errors and dynamics discretization errors.
Solution accuracy is progressively improved by varying the number of ADMM iterations at termination from 20 to 200 iterations.
The inertial modeling errors are introduced in the simulated robot by randomly varying the mass, center of mass location and moments of inertia of each link by $\pm 10, 20, 30 \mathrm{~and~}40 \%$.
Dynamics discretization errors are introduced in the NMPC by increasing the integration timestep $dt$ from 20~ms to 50~ms while keeping a constant time horizon of 0.8~s.
A larger $dt$ reduces the number of decision variables and solve times at the expense of reduced model fidelity.
The simulations are performed in parallel using the MIT Supercloud High Performance Computing cluster \cite{reutherInteractiveSupercomputing402018}.
Finally, we add the CBFs and compare their effect for $dt=25$~ms.
\par
For each controller tested, we analyse the number of successful simulations, normalized by the maximum number of successes obtained by the most successful controller (i.e. \new{255 successes} for $dt=20$~ms at 150 ADMM iterations, without inertial modeling error).

\subsection{Effect of Inertial Modeling Errors}
\label{sec:inertial_error}
\begin{figure}[!htbp]
    \centering
    \includegraphics[scale=1]{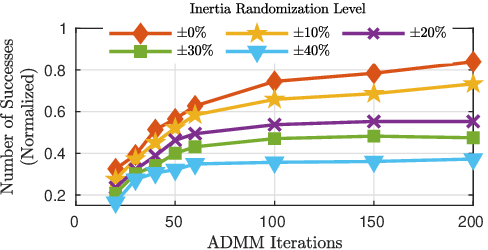}
    \caption{Normalized number of successes for increasing ADMM iterations and inertial modeling errors without CBFs and computation delay \new{for $dt$ = 25~ms}. \new{The potential success improvement decreases as the inertia randomization increases (i.e., 51, 46, 32, 27 and 21\%, respectively).}}
    \label{fig:inertial_error_effect}
\end{figure}

Fig. \ref{fig:inertial_error_effect} shows the number of successful simulations against the number of ADMM iterations for increasing inertial modeling errors, \new{for $dt=25$~ms}.
The number of successes increases with the number of ADMM iterations.
\par
However, inertial modeling errors render higher-accuracy solutions less effective.
We computed the difference between the number of successes at 20 ADMM iterations and the maximum number of successes for each level of inertial model randomization in Fig. \ref{fig:inertial_error_effect}.
This assesses the potential success improvement obtainable by increasing solution accuracy when computation delay is neglected.
The potential for success improvement respectively reduces from 51\% to 46, 32, 27 and 21\% for inertial errors of $\pm$ 0, 10, 20, 30 and 40\%, indicating that the real MIT Humanoid may not benefit significantly from higher solution accuracy.
These effects should be more pronounced on hardware, since our simulator does not model foot rolling contact, backlash, transmission compliance, sensor noise, or low-level current control.

\subsection{Effect of Dynamics Discretization Error}
\label{sec:discretization_error}
\begin{figure}[!htbp]
    \centering
    \includegraphics[scale=1]{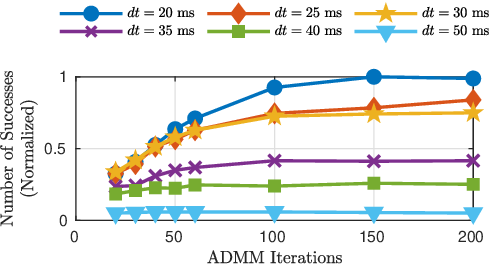}
    \caption{Normalized number of successes  for increasing ADMM iterations and integration timestep without CBFs and computation delay. \new{The potential success improvement decreases as the integration timestep increases (i.e., 67, 51, 41, 18, 7 and 1\%, respectively).}}
    \label{fig:dt_error_effect}
    \vspace{-4mm}
\end{figure}

Fig. \ref{fig:dt_error_effect} shows the normalized number of successful simulations against the number of ADMM iterations for increasing discretization timesteps.
Between 20 to 40 iterations, the controllers with $dt \in \{20, 25, 30\}$~ms perform identically, and it takes many more iterations for $dt=20$~ms to outperform the coarser timesteps.
\par
The potential success improvement respectively decreases from 67\% to 51, 41, 18, 7 and 1\% as timesteps increase from 20~ms to 25, 30, 35, 40 and 50~ms, indicating that models with larger discretization errors, typically used for real-time NMPC, may not benefit as much from increased solution accuracy when used in a high-fidelity simulator and on hardware.
For instance, the controller with $dt=50$~ms does not benefit from running more than 20 ADMM iterations.
Despite being the least successful, it made the real robot walk at up to 0.8~m/s.

\subsection{Effect of Computation Delay}
\label{sec:delay:effect}
\begin{figure}[!htbp]
    \centering
    \includegraphics[scale=1]{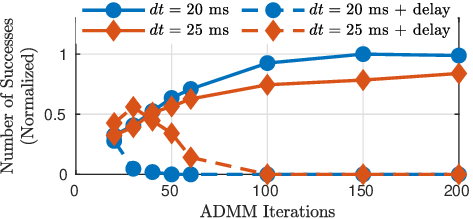}
    \caption{Normalized number of successes for increasing ADMM iterations with and without computation delay for $dt \in \{20, 25\}$~ms.}
    \label{fig:success_vs_delay}
    \vspace{-5mm}
\end{figure}

Even though controllers with accurate models benefit the most from accurate solutions, the longer computation times offset these benefits.
Fig. \ref{fig:success_vs_delay} illustrates the impact of delay on the number of successes for timesteps of $dt \in \{20, 25\}$~ms.
The formulation with $dt = 20$~ms is too slow for real-time use due to its longer horizon ($N=54$).
The delay renders higher accuracy solutions ineffective despite using prediction delay compensation.
Conversely, for $dt = 25$~ms and above, the computation time is sufficiently short to benefit from increased ADMM iterations.
Specifically, for $dt=25$~ms, the number of successes increases 
from 20 to 30 ADMM iterations but declines beyond that point.
\par
\new{
    Surprisingly, the number of successes is 17\% higher with delay for 30 ADMM iterations.
    We found that the low-level PD control law \eqref{eq:pd_ctrl} becomes more effective under delay when paired with the prediction delay compensation strategy.
    Without delay, the feedback terms have minimal impact, because the setpoints from the frequently updated trajectory remain close to the current state.
    In contrast, with delay, the interpolation strategy causes the setpoints to be farther from the current state, thereby enhancing the feedback effect.
    }
\par
\new{We also investigated the effect of a second SQP iteration for $dt=25$~ms.
Without delay, the number of successes increased by 13\% at 30 ADMM iterations.
However, delay makes the second SQP iteration ineffective, as success dropped to 0.02.}

\subsection{Effect of Control Barrier Functions}
\begin{figure}[!htbp]
    \centering
    \includegraphics[scale=1]{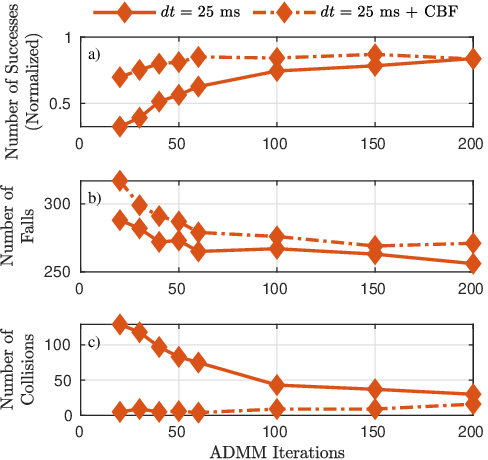}
    \caption{Effect of the CBFs on a) the normalized number of successes, b) the number of falls and c) the number of self-collisions for $dt=25$~ms.}
    \label{fig:cbf_comp}
\end{figure}

Fig. \ref{fig:cbf_comp} illustrates the number of successes (a), falls (b), and self-collisions (c) for $dt = 25$~ms, with and without CBFs.
Without CBFs, the solutions from ADMM yield a small number of falls within 12\% of the minimum number of falls at 20 iterations, but the number of collisions reduces by 4.3 times from 20 to 200 iterations.
\par
The CBF constraints at the first timestep are independent of the initial guess and, therefore, easier to satisfy.
As a result, they almost eliminate the self-collisions, regardless of the number of ADMM iterations.
The number of falls increases by up to 10\% since CBFs prevent infeasible motions such as penetrating legs.
Overall, the number of successes doubles at 20 ADMM iterations and matches the success level obtained at 100 iterations without CBFs.

\par
\begin{figure}[!htbp]
    \centering
    \includegraphics[scale=1]{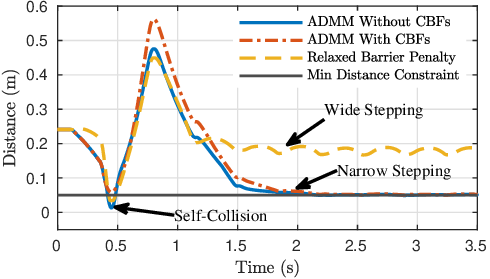}
    \caption{Distance constraint between the feet for a lateral push recovery simulation. Our proposed approach (ADMM with CBF) is compared against ADMM without CBFs and the relaxed-barrier penalty method \cite{grandiaPerceptiveLocomotionNonlinear2023}. The disturbance is applied at 0.35~sec.}
    \label{fig:lat_push_comparison}
    \vspace{-5mm}
\end{figure}

We further illustrate the effect of CBFs in a lateral push experiment, where the robot recovers from a leftward force applied to the torso while the left leg is in stance and the right leg is in swing.
This scenario is challenging because the right swing leg must step as close as possible to the left stance leg without colliding to avoid falling.
Fig. \ref{fig:lat_push_comparison} shows the distance constraint between the feet for different strategies: ADMM without CBFs, ADMM with CBFs, and the relaxed-barrier penalty method from \cite{grandiaPerceptiveLocomotionNonlinear2023} for a disturbance applied at 0.35~s.
For ADMM without CBFs, the distance constraint is violated by 3.7~cm at 0.45~s.
However, sufficient SQP iterations allow the robot to recover with feasible narrow stepping that minimizes lateral torso velocity after 2~s.
The penalty method achieves lower constraint violation (1.6~cm), but couples cost minimization and constraint satisfaction, leading to suboptimal wider steps after 1.5~s.
Our approach, which combines ADMM and CBFs, maintains strict constraint satisfaction without altering the optimal narrow stepping behavior after recovery.

\section{Hardware Experiments}
\label{sec:hardware}
The hardware results are best captured in the supplementary video linked in Fig. \ref{fig:hardware_figures}.
The NMPC allows the MIT Humanoid to holistically reason about all degrees of freedom while respecting joint limits and self-collision constraints with sufficient accuracy.
This enables the robot to coordinate its arm and leg motions to dynamically balance when walking, stopping, and recovering from significant disturbances (Fig. \ref{fig:hardware_figures}).
In practice, joint position limits are always respected, allowing the robot to walk stably with its knees fully extended.
With the sphere-based collision model from Fig. \ref{fig:self_collision_constraint}b and CBFs, the robot reliably plans complex crossed-leg motions that enhance stability in challenging push recovery, abrupt halting, or fast turning scenarios.
\par
The different NMPC formulations transfer seamlessly to hardware.
Table \ref{table:computation_breakdown} shows the solve times of four NMPC formulations extensively tested on hardware with 20 ADMM iterations.
The full collision model (Fig. \ref{fig:self_collision_constraint}b) and the CBFs only increases solve times by 9\% and 2.7\%, respectively.
Note that the standard deviations are low due to the fixed number of iterations.
The formulation with $dt=50$~ms is the fastest one that remains stable on hardware and can walk at up to 0.8~m/s at a 200~Hz update rate.
The formulation with $dt=25$~ms is the most successful controller (Fig. \ref{fig:dt_error_effect}) that is still fast enough for online use (90~Hz).
It can walk at speeds of up to 1.1~m/s on hardware and is much more resistant to push recoveries and rough terrain.

\begin{table}[!htbp]
    \centering
    \caption{\centering Computation Time Breakdown on the MIT Humanoid's Computer for 20 ADMM Iterations While Walking }
    \label{table:computation_breakdown}
    \begin{tabular}{|cc|cccc|}
    \hline
    \multicolumn{2}{|c|}{\begin{tabular}[c]{@{}c@{}}NMPC\\ Formulation\end{tabular}}                                                      & \multicolumn{4}{c|}{\begin{tabular}[c]{@{}c@{}}Computation Time\\ (ms)\end{tabular}}                                                                                                                                                                                                      \\ \hline
    \multicolumn{1}{|c|}{\begin{tabular}[c]{@{}c@{}}$dt$\\ (ms)\end{tabular}} & \begin{tabular}[c]{@{}c@{}}Collision\\ Model\end{tabular} & \multicolumn{1}{c|}{\begin{tabular}[c]{@{}c@{}}QP\\ Data\end{tabular}} & \multicolumn{1}{c|}{\begin{tabular}[c]{@{}c@{}}QP\\ Step\end{tabular}} & \multicolumn{1}{c|}{\begin{tabular}[c]{@{}c@{}}Linesearch\end{tabular}} & \begin{tabular}[c]{@{}c@{}}Total\\ mean (std)\end{tabular} \\ \hline
    \multicolumn{1}{|c|}{50}                                                  & Simple                                                    & \multicolumn{1}{c|}{$0.9$}                                             & \multicolumn{1}{c|}{$3.3$}                                             & \multicolumn{1}{c|}{$0.13$}                                                & $4.3~(0.2)$                                                \\ \hline
    \multicolumn{1}{|c|}{50}                                                  & Full                                                      & \multicolumn{1}{c|}{$1.0$}                                             & \multicolumn{1}{c|}{$3.5$}                                             & \multicolumn{1}{c|}{$0.15$}                                                & $4.7~(0.2)$                                                \\ \hline
    \multicolumn{1}{|c|}{25}                                                  & Full                                                      & \multicolumn{1}{c|}{$2.8$}                                             & \multicolumn{1}{c|}{$7.8$}                                             & \multicolumn{1}{c|}{$0.4$}                                                 & $11.0~(0.2)$                                               \\ \hline
    \multicolumn{1}{|c|}{25}                                                  & Full + CBF                                                & \multicolumn{1}{c|}{$2.9$}                                             & \multicolumn{1}{c|}{$7.9$}                                             & \multicolumn{1}{c|}{$0.5$}                                                 & $11.3~(0.2)$                                               \\ \hline
    \end{tabular}
    \vspace{-5mm}
\end{table}
\section{Conclusion and Outlook}
\label{sec:conclusion}
This paper presented an implementation of whole-body NMPC that plans trajectories for all the degrees of freedom of the MIT Humanoid in real time.
Our simulation results suggest that real robots may not benefit from high-accuracy solutions due to unavoidable inertial modeling errors, dynamics discretization errors and computation delay.
Instead of seeking high accuracy, we leverage ADMM's ability to provide reliable, low-accuracy \newtwo{QP subproblem} solutions at a high frequency and show their adequacy through extensive hardware experiments.
Furthermore, we use CBFs at the first timestep of the NMPC to enhance the closed-loop feasibility of self-collision constraints with minimal added computational burden.
With our approach, the MIT Humanoid reliably walks and responds to significant disturbances with holistic whole-body reasoning while respecting contact constraints, joint limits, and self-collisions.
This allows the robot to execute arm and crossed-leg motions that enhance stability.
\par
Despite its reliability, our approach cannot guarantee stability or feasibility of the solution.
Additionally, the controller follows a fixed contact sequence.
This makes the robot vulnerable to large pushes applied at the beginning of a swing phase, \new{where the time until the next foot touchdown is maximal.}
Finally, the most successful controllers in simulation are still too slow for hardware deployment.
Our solver can be accelerated by state-of-the-art advancements such as analytical derivatives \cite{carpentierAnalyticalDerivativesRigid2018}, parallelization \cite{katayamaEfficientSolutionMethod2021}, and Riccati factorizations \cite{jordanaStagewiseImplementationsSequential2023}.
\section*{Acknowledgments}
We thank Andrew SaLoutos, Hongmin Kim and David Nguyen for continuously robustifying the MIT Humanoid hardware.





\small\printbibliography

\end{document}